\documentclass{article}
\advance \topmargin by -1.0in
\advance \oddsidemargin by -1in

\usepackage{url}

\usepackage{longtable}

\usepackage{booktabs}

\makeatletter
\def\set@curr@file#1{\def\@curr@file{#1}} 
\makeatother

\usepackage{booktabs}
\usepackage{times}
\usepackage{latexsym}

\usepackage{booktabs}
\usepackage{stmaryrd}
\usepackage{graphicx,color}
\usepackage{amssymb}
\usepackage{amsmath}
\usepackage{xspace}
\usepackage[font=footnotesize]{caption}
\usepackage{tikz}
\usepackage{pgfplots}
\usepackage{authblk}

\usepackage[utf8]{inputenc}

\def\para#1{\vspace{0.1in}\noindent {\bf #1}}

\pgfplotsset{compat=1.17}

\title{Improving Early Sepsis Prediction with Multi Modal Learning}

\author[1]{Fred Qin\thanks{fqin22@cmc.edu}}
\author[2]{Vivek Madan\thanks{vivmadan@amazon.com}}
\author[3]{Ujjwal Ratan\thanks{ujjwalr@amazon.com}}
\author[4]{Zohar Karnin\thanks{zkarnin@amazon.com}}
\author[5]{Vishaal Kapoor\thanks{vishaalk@amazon.com}}
\author[6]{Parminder Bhatia\thanks{parmib@amazon.com}}
\author[7]{Taha Kass-Hout\thanks{tahak@amazon.com}}
\affil[1]{Claremont McKenna College}
\affil[2,3,4,5,6,7]{AWS AI Labs}
\date{}

\begin{document}

\maketitle

\begin{abstract}
Sepsis is a life-threatening disease with high morbidity, mortality and healthcare costs. The early prediction and administration of antibiotics and intravenous fluids is considered crucial for the treatment of sepsis and can save potentially millions of lives and billions in health care costs. Professional clinical care practitioners have proposed clinical criterion which aid in early detection of sepsis; however, performance of these criterion is often limited. 
Clinical text provides essential information to estimate the severity of the sepsis in addition to structured clinical data. In this study, we
explore how clinical text can complement structured data towards early sepsis prediction task. In this paper, we propose multi modal model which incorporates both structured data in the form of patient measurements as well as textual notes on the patient. We employ state-of-the-art NLP models such as BERT and a highly specialized NLP model in Amazon Comprehend Medical to represent the text. On the MIMIC-III dataset containing records of ICU admissions, we show that by using these notes, one achieves an improvement of $6.07$ points in a standard utility score for Sepsis prediction and $2.89\%$ in AUROC score. Our methods significantly outperforms a clinical criteria suggested by experts, qSOFA, as well as the winning model of the PhysioNet Computing in Cardiology Challenge for predicting Sepsis. \cite{SepsisChallenge}.

\end{abstract}

\section{Introduction}

Sepsis is an extreme body response to an infection that occurs when an infection in the body triggers a chain reaction throughout the body. It can lead to tissue damage, organ failure and death. Despite
promising medical advances over the last decades, sepsis remains among the most common
causes of in-hospital deaths. It is associated with an alarmingly high mortality and morbidity,
and massively burdens the health care systems world-wide. In the United States alone, sepsis causes 270,000 deaths per year which accounts for over one third of total U.S. hospital related deaths annually.  Cost of sepsis management in U.S. hospitals annually is around \$24 billion and majority of these costs occur for patients who developed sepsis during the hospital stay. This exceed the cost for any other health condition. Also, studies have shown that mortality rate increase by 4-8\% 
if there is a delay in Sepsis identification by just one hour~\cite{SeymourGPFIPLOTL17,LiuFGBIBE17}. The early prediction and administration of antibiotics and intravenous fluids is considered crucial for the treatment of sepsis and can save potentially millions of lives and billions in health care costs.

A broad variety of disorders fall in the umbrella of sepsis, severe sepsis and septic shocks. These are characterized by a dysregulated response from a host to an infectious insult. However, due to the heterogeneous nature of infections as well as the host response, these disorders have been difficult to recognize by the physicians. A redefinition of sepsis, sepsis-3 has been introduced with the aim of increasing correct identification of sepsis in clinical and preclinical settings. It eliminates the traditional classification from sepsis, through severe sepsis and septic shock. Instead, it introduces a two-tier identification system that is tied to increase in mortality probability. It defines sepsis as a life-threatening organ dysfunction caused by a dysregulated host response to infection.  

Early prediction of sepsis remains challenging because sepsis exhibits similar symptoms as less critical conditions, especially in intensive care units \cite{GohWYPLYT21}.  There have been a series of sepsis identification tools suggested by critical care professionals: (i) sequential organ failure assessment (SOFA), an organ failure assessment system including 7 criteria across 6 organ systems used in the Sepsis-3 definition, (ii) systematic inflammatory response syndrome (SIRS), an inflammatory response score containing 4 criteria used in previous definitions of sepsis, (iii) logistic organ dysfunction system (LODS), an organ failure assessment score taking into account 6 organ systems that is more complex and less well known than SOFA, and (iv) quick sequential organ failure assessment (qSOFA), a simple scoring system including 3 criteria that was designed to quickly assess sepsis at the bedside without lab tests \cite{SingerDSSABBBCC16, jones2009sequential, le1996logistic}. There are other similar medical scores that are applied more generally such as the modified early warning score (MEWS) which is supposed to indicate when a patient is entering a critical state and simplified acute physiology score (SAPS II) which attempts to assess probability of mortality \cite{gardner2006value, le1993new}. All of these methods can be applied for early prediction of Sepsis. However, they suffer from low performance when predicting sepsis early and often require time-consuming laboratory results~\cite{IslamNWWYL19,DesautelsCHJKSSCFB16}. These methods all make use of various vital signs and laboratory results to generate the sepsis risk scores but fail to analyze trends in patient data or correlation between measurements. 


There have been some previous work around using machine learning at solving this problem with some success. For instance, InSight build a machine learning model based on the seven vital signs available at the bedside to predict if the patient is likely to develop sepsis in the near future. It does not use any of the laboratory tests and still significantly outperforms SIRS, qSOFA, MEWS, SAPS II and SOFA method~\cite{DesautelsCHJKSSCFB16}. Furthermore, the PhysioNet Computing in Cardiology Challenge 2019 \cite{SepsisChallenge} and 2019 DII National Data Science Challenge have provided data sets for early sepsis prediction that have led to dozens of approaches to sepsis prediction including LSTM-based models \cite{zhang2021interpretable}, random forest classification \cite{lyra2019early}, and gradient boosting methods \cite{YangWGLLL19}. Though, these works show improvement in performance for early sepsis prediction, they focus on building predictive models using only the structured part of the EHR. These solutions have ignored the large amount of information present in clinical text including doctor notes, discharge summaries and other available textual features. In this paper, we present an architecture incorporating textual features in addition to the structured numerical features. Formally, we attempt to solve the following problem:
\begin{center}
{\em Build a predictive model which based on numerical as well as textual features predicts with high confidence if the patient will develop Sepsis in the near future.}
\end{center}

We explore two different methods of incorporating text into our model. First approach is about  utilizing large pretrained language models.
A language model is pre-trained on a large amount of unstructured text corpus such as Wikipedia, bookcorpus etc., and learns contextualized embeddings for any piece of text. An embedding is a numeric representation of the text that can be processed by standard machine learning algorithms. For a classification task involving textual input, it can be used to build a predictive model by training a classifier on the contextualized embedding given by the language model on the task's textual data. Out of the many language models proposed in NLP, BERT stands out~\cite{DevlinCLT18}. It performs extremely well and achieves state-of-the-art performance for virtually all NLP tasks. In our model, we use the contextualized embeddings given by BERT when applied on the clinical notes along with numerical features such as blood pressure, heart rate etc., to train a classifier for predicting sepsis.  We use a specialized BERT(ClinicalBERT), pre-trained on clinical notes~\cite{HAR19}. Our second method involves extrating clinical entities such as medications, medical conditions, tests, treatments and procedures from unstructured clinical text through  Amazon's Comprehend Medical~\cite{comprehendmedical}
. This service extracts entities which can be more relevant to predicting sepsis than the entire text. We incorporate these entities into our model by computing either a BERT embedding or a Term Frequency Inverse Document Frequency (tf-idf) embedding as a way to represent text in our model.


We evaluate the performance of our model on the MIMIC-III dataset consisting of patient records from roughly $60,000$ intensive care unit (ICU) admissions~\cite{JohnsonPSLFGMSCM16}. It is compiled from the Beth Israel Deaconess Medical Center (BIDMC) in Boston, MA between 2001 and 2012. It includes demographics, vital signs, laboratory tests, medications, nursing/clinician notes and more. It is the most widely used dataset for clinical NLP research.
To the best of our knowledge, this is the first
work to jointly model named entity and structured features for Sepsis in an end-to-end system. Our techniques for incorporating text can be applied for any of the related problems such as mortality prediction, morbidity prediction etc. Our main contributions
are summarized below:
\begin{itemize}
    \setlength\itemsep{0.02em}
    \item We build models for the most up-to-date definition of sepsis (Sepsis-3) and show several important data processing details for MIMIC-III.
    \item We extend the current models to incorporate text into the sepsis prediction models.
    \item We explore two different methods based on the state-of-the-art language model, BERT and an entity extraction model specialized for clinical text, Comprehend Medical. 
    \item We perform extensive analysis of the proposed models.  Our method of incorporating text via BERT embeddings improve AUROC score by $2.89\%$ and utility score by $6.07$ points over the competing method not utilizing text \cite{YangWGLLL19} which won the PhysioNet Computing in Cardiology Chellenge for predicting Sepsis \cite{SepsisChallenge}. It corresponds to a decrease in false sepsis prediction rate by $4.09\%$ and an increase in true sepsis prediction rate by more than $4\%$ upto $3$ hours before the sepsis onset time. Our method also outperforms the baseline methods qSOFA, a clinical criteria suggested by experts to quickly asses Sepsis without lab tests and a method utilizing tf-idf embeddings to incorporate text into the model.
\end{itemize}

\section{Textual Features} \label{sec:text}
Clinical text provides essential information to estimate the severity of the sepsis in addition to structured clinical data.   In this study, we explore how clinical text can complement structured data towards early sepsis prediction task. Most of the prior work on sepsis prediction have used the structured part of the EHR.  However, there are still a lot of multitype data in ICU database that have not been utilized effectively, especially clinical notes. In fact, in the actual medical diagnosis process, the contents involved in clinical notes are also important information that doctors need to measure. Furthermore, the history of the illness and other historical information should be considered for patients with chronic diseases, but not for people with non-chronic diseases. There could be important signals in these notes for predicting sepsis. 
The main focus of this paper is in addressing this issue and building models which incorporate textual features. The study of textual representation and modeling is the field of natural language processing (NLP). We borrow these techniques from NLP for textual representation and apply it to sepsis prediction models. 
\para{Term frequency inverse document frequency (tf-idf)} is a classic method of embedding text information into a numeric vector. An input document (e.g.\ doctor's notes) is converted into a set of words in a pre-determined vocabulary (of say 1000 words). For every vocabulary word we compute the number of documents in which it appears (document frequency). For every document, the resulting embedding has an entry for each vocabulary word. Its value is the number of times the word appears in the document times a term dependant of the inverse document frequency, aimed to penalize words that are too frequent. This technique is commonly used to embed text. Its greatest disadvantage is that it does not provide external knowledge, as opposed to the methods reviewed next.

\para{BERT-based embeddings:} Pretrained language models such as such as ELMo, GPT,  Transformer-xl, BERT, and XLNet
have become a norm achieving state of the art performance across most of the natural language processing tasks. Bidirectional Encoder Representation from Transformers (BERT) stands out among these models and has become a key component in solving virtually all natural language tasks~\cite{DevlinCLT18}. A standard pipeline for using such models is the following: (i) Pre-train the model on a large corpus such as wikipedia such that it can learn meaningful contextual embedding for every piece of text. An embedding is a numeric representation of the text that can be processed by a standard machine learning algorithm. (ii) For classification tasks, train a classification model on the contextualized embeddings computed by the pre-trained model for the task data. Embeddings are often fine tuned as well in the second step, meaning the embedding function is slightly modified in order to improve the classification quality. In our model, we use the contextualized embeddings from the pre-trained model on the textual features for the patient and concatenate it to the numerical features. The resulting list of features are fed into the classification algorithm. Since there are multiple textual features, there are two ways of getting contextualized embeddings: (i) merge all the textual features and get contextualized embedding ({\bf BERTM}), (ii) get contextualized embeddings separately for each textual feature and concatenate all the embeddings ({\bf BERTS}). We explore both of these options.
Instead of using BERT pre-trained on a large general domain corpus, we use the embeddings generated by ClinicalBERT, a BERT model pre-trained on clinical notes~\cite{HuangAR19}. Embeddings produced by ClinicalBERT are of higher quality than BERT for our setting as it is trained on text corpus from clinical domain; this has been pointed out in previous works as well. For ease of convenience, we refer to ClinicalBERT as BERT in the subsequent discussion.

In addition to the embeddings generated by the pre-trained BERT, we explore task specific training. Here we use the probability score from a finetuned BERT as textual representation. We first fine-tune the BERT model on the textual features for the task of Sepsis Prediction. This involves fine tuning the embeddings as well as training a linear classifier on the embeddings to predict Sepsis. In this phase, we ignore the numerical features for the patients. Once we get a BERT based classification model, we use the probability score of the text given by this fine tuned model as the textual representation in our pipeline. To avoid information leakage, both the BERT fine tuning as well as training of our model is done on the train set which is disjoint from the test set where we evaluate the performance of our model. When discussing our experiments we refer to this technique as either {\bf FBERTM} or {\bf FBERTS}, depending on whether we merge all the textual features and generate a single probability score or generate a probability score for each textual feature.



\para{Amazon Comprehend Medical:} Amazon Comprehend Medical is a fully-managed service that performs named entity recognition on unstructured clinical notes. It uses a pre-trained model to analyze unstructured clinical text through entity detection. An entity is a textual reference to medical information such as medical conditions, medications, or Protected Health Information (PHI). In addition, it can also derive RxNorm codes for medications and  ICD-10 codes for medical conditions.
The MIMIC-III data set contains unstructured text that contains multiple clinical entities about the patient. Amazon Comprehend medical can extract these entities which can then be correlated with the structured portions of the patient records. These clinical entities can be included in the feature set of the patients as a list. It has been demonstrated that the inclusion of these features provide richer information embedded within unstructured clinical notes to the algorithm and helps it to perform better~\cite{comprehendmedicalpost}. We incorporate this into our model by first computing the entities in the textual features for a patient and computing a tfidf embedding ({\bf CMtfidf}) or a  contextualized embedding using BERT ({\bf CMBERT}).

\begin{figure}[ht]
    \centering
    \includegraphics[scale=0.35]{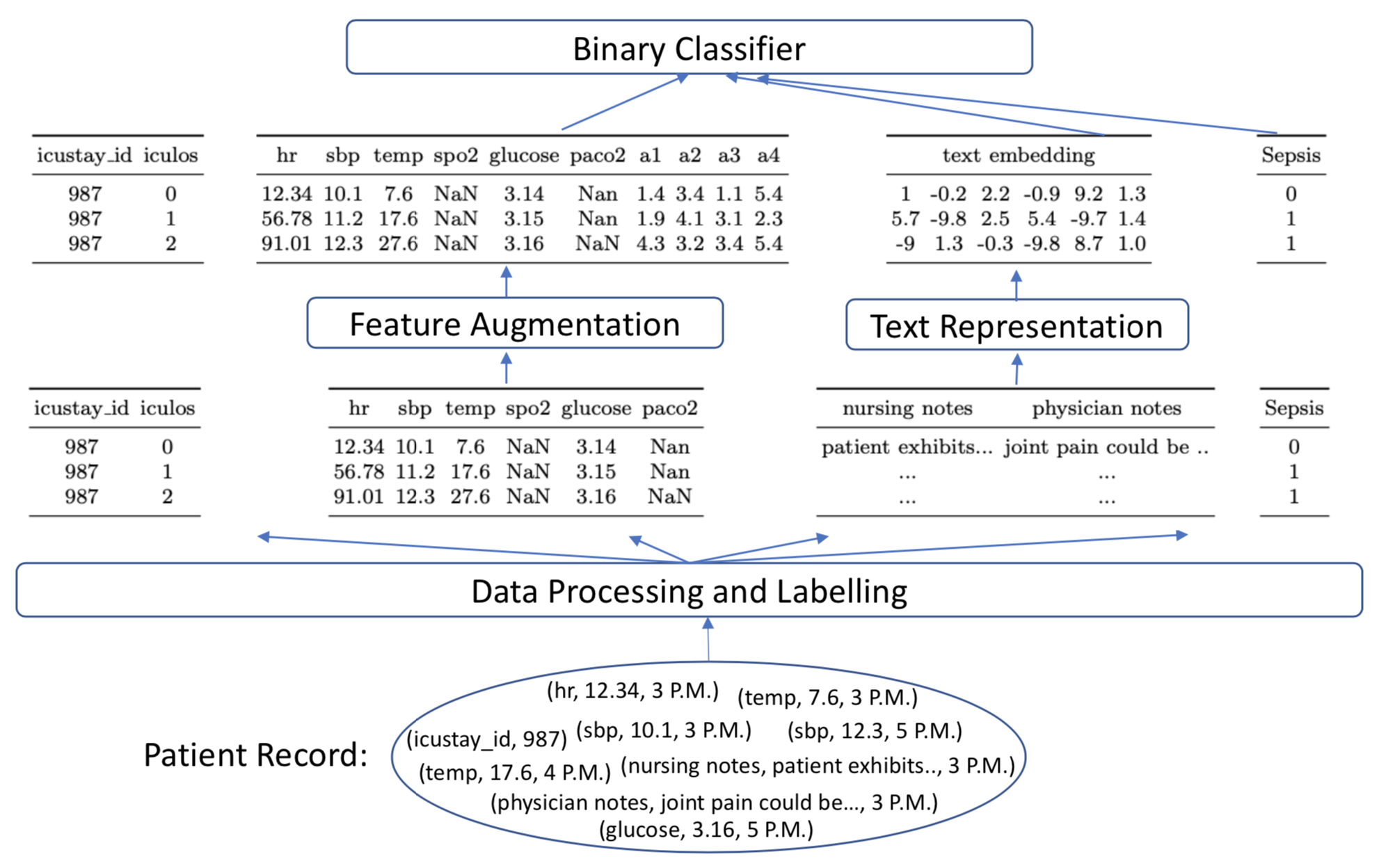}
    \caption{Training pipeline
    }
    \label{fig:alg_pipeline}
\end{figure}

\section{Dataset and the training pipeline}
MIMIC-III is an openly available dataset developed by the MIT Lab for Computational Physiology, comprising deidentified health data associated with roughly 60,000 intensive care unit (ICU) admissions~\cite{JohnsonPSLFGMSCM16}. It includes demographics, vital signs, laboratory tests, medications, and more. It is compiled from the Beth Israel Deaconess Medical Center (BIDMC) in Boston, MA between 2001 and 2012. Each observation is associated with a patient id, feature name, feature value and the time of observation.

\subsection{Determining Sepsis-3 onset}\label{subsec:sepsis_definition}
Although sepsis is a prevalent cause of medical concern for hospitals, the exact definition, especially regarding the time of exact sepsis onset has evolved and has been a topic for debate in the medical community. Fortunately, a team of experts rigorously defined sepsis in 2016, leading to the Third International Consensus~\cite{JonesTK09}. This was the third and most recent iteration of defining sepsis. In 2019, PhysioNet challenge slightly perturbed the definition to use a more restrictive criteria~\cite{SepsisChallenge}. We will be using this more restrictive definition of Sepsis-3 in our experiments. 
As per this definition, there are two components of a patient that has sepsis:
\begin{itemize}
    \item {\em Suspicion of infection.}
Clinical suspicion of infection identified as the earlier timestamp of IV antibiotics and blood cultures within a specified duration.
If antibiotics were given first, then the cultures must have been obtained within 24 hours. If cultures were obtained first, then antibiotic must have been subsequently ordered within 72 hours.
Antibiotics must have been administered for at least 72 consecutive hours to be considered.
\item  {\em SOFA.}
The SOFA (Sequential Organ Failure Assessment) score was a score established by a paper in 2009 that numerically quantifies the number and severity of failed organs
The occurrence of end organ damage as identified by a two-point deterioration in SOFA score within a 24-hour period. The time that this occurred is defined by the first collection of data that leads to a measurement causing a two-point deterioration
\end{itemize}

The time of onset for sepsis is defined as the timestamp of the earliest time of these two components if they are within a 24-hour period. If one of these events does not occur or they are not within 24 hours, the patient is defined as negative for sepsis.

\subsection{Feature Selection}\label{sec:feature_selection}
There are several  possible features available for patients. 
We selected a subset of these features based on their frequency and their relevance. Features that are too frequently missing are not used.
%
%
The first set of features include vital signs which are taken at frequent intervals and widely used for all medical prediction tasks~\cite{SepsisChallenge,ScherpfGMZ19}. These include (i) heart rate (hr, beats per minute), (ii) scholastic blood pressure (sbp, in millimeters of mercury), (iii) diastolic blood pressure (dbp, millimeters of mercury) (iv) mean arterial pressure (map, millimeters of mercury), (v) respiratory rate (resp, breaths per minute) and (v) temp (temperature, degree Celsius). 
Similar to prior work~\cite{ScherpfGMZ19}, we chose a few more features based on being widely available and relevant to the prediction of Sepsis including:
(vii) oxygen saturation in \% using pulse oximetry (spo2) (viii) blood glucose levels (glucose, milligrams per decilitre) and (ix) partial pressure of carbon dioxide (paco2, millimeters of mercury).

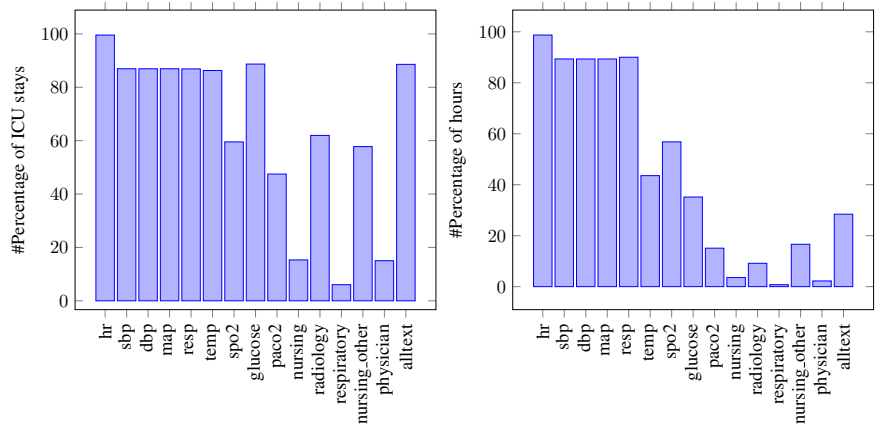
\begin{figure}
    \centering
\begin{tikzpicture}[scale=0.7]
\begin{axis}[  
    ybar,  
    ylabel={\#Percentage of ICU stays}, 
    xticklabel style={rotate=90},
    symbolic x coords={ hr, sbp, dbp, map, resp, temp, spo2, glucose, paco2, nursing, radiology, respiratory, nursing\_other,  physician, alltext}, 
    xtick=data,  
    nodes near coords align={vertical},  
    ]  
\addplot coordinates {(hr, 99.56) (sbp, 86.94) (dbp, 86.93) (map, 86.94) (resp, 86.88) (temp, 86.28) (spo2, 59.54) (glucose, 88.69) (paco2, 47.49) (nursing,15.32) (radiology, 61.96) (respiratory,6.04) (physician, 15.02) (nursing\_other,57.78) (alltext,88.59)};  
\end{axis}
\end{tikzpicture}
\begin{tikzpicture}[scale=0.7]
\begin{axis}[  
    ybar,  
    ylabel={\#Percentage of hours}, 
    xticklabel style={rotate=90},
    symbolic x coords={ hr, sbp, dbp, map, resp, temp, spo2, glucose, paco2, nursing, radiology, respiratory, nursing\_other, physician,alltext}, 
    xtick=data,  
    nodes near coords align={vertical},  
    ]  
\addplot coordinates {(hr, 98.73) (sbp, 89.35) (dbp, 89.33) (map, 89.34) (resp, 90.02) (temp, 43.56) (spo2, 56.82) (glucose, 35.17) (paco2, 15.10) (nursing,3.56) (radiology, 9.17) (respiratory,0.77) (physician, 2.24) (nursing\_other,16.63) (alltext,28.43)};  
\end{axis}
\end{tikzpicture}
\caption{Figure shows the percentage of patients with at least one record of a specific feature. alltext corresponds to the percentage of ICU stays/hours with at least one of the five textual features.}
\label{fig:histogram}
\end{figure}

In addition to numerical features, we have access to several textual features available for patients. These include nursing notes, physician notes, radiology notes, respiratory notes, case management notes, consult notes, discharge summary, ECG notes, echo notes, general notes, nutrition notes, pharmacy notes, rehab services notes and social work notes. We only use nursing notes, physician notes, radiology notes and respiratory notes as rest of the features except discharge summary are available for insignificant percentage of patients. Discharge summary is not useful for sepsis prediction as it occurs at the end of the ICU stay. There are two types of nursing notes, nursing and nursing\_other. Nursing are typically long notes whereas nursing\_other are quick and more frequent notes. We use both of these in our model.  Statistics on the number of patients and the number of hours containing each of the features can be found in Figure~\ref{fig:histogram}.

\subsection{Data Processing and Labelling}\label{sec:data_processing_labelling}
MIMIC-III contains information related to $61532$ patients admitted to ICU. 
Out of these patients, $2145$ patients have no relevant features available. Thus, we ignore these patients as we can not hope to build a model which can predict Sepsis. As per definition discussed in Section~\ref{subsec:sepsis_definition}, $12210$ have Sepsis. However, $8082$ of the patients are admitted to the ICU 
with Sepsis and $2137$ develop sepsis within the first 6 hours in the ICU. For such patients, we have either no data or very little data before the sepsis onset. So, we can not hope to build a model which can predict Sepsis in advance for these patients. Thus, we ignore these patients as well. For the resulting $49168$ patients, $1991$ ($4.05\%$) patients have Sepsis. For these patients, we have a total of $1027079$ hours of data available with an average of roughly $21$ hours of ICU stay data per patient. We create a data point for each hour of ICU stay for each patient. Features include the measurements for the patient in that hour. If multiple measurements are available for single hour, we take the average value for numeric values, and concatenate for text.


Our machine-learning training procedure involves training a binary classifier. This procedure requires positive examples, meaning (patient,hour) pairs that should be predicted as a patient that will develop Sepsis, and negative example, meaning (patient,hour) pairs that should be predicted as a patient with no risk for Sepsis. To obtain negative examples, we take all the (patient, hour) pairs of patients that did not develop Sepsis. To obtain positive examples we take all (patient, hour) pairs of septic patients whose Sepsis onset time is between 0 to 6 hours after the given hour. The decision to use this particular definition of positive examples is done based on a similar decision made in the PhysioNet Computing and Cardiology challenge for predicting Sepsis \cite{SepsisChallenge}. Although it affects the training procedure, the evaluation procedure (detailed in Section~\ref{sec:utility_score}) is independent of this decision. 


\subsection{Feature Augmentation}
In the data processing step above, we construct a datapoint containing observations for all hours of all patients. This however ignores some important information of the data. Data could be missing in a specific hour. Also, history and progression of features such as blood pressure or temperature may also have a signal for predicting Sepsis which is not present in the most recent measurement of these features. To incorporate the missing features, we forward fill the features; use the measurement from the most recent hour if a measurement is unavailable. To incorporate the progression of features, for each of the nine numerical features discussed in Section~\ref{sec:feature_selection} we first add three differential/missingness features: (i) Number of hours since the last available value of this feature (If it was never observed, we set it to -1, and if the feature is currently present, it is set to 0) (ii) number of available values of this feature in the hours that proceeded this one and (iii) Difference in the value of the feature for this datapoint and the datapoint in the previous hour.  Furthermore, for each of the five most common features hr, sbp, map, debp and resp, we add six sliding window features: min, max, mean, median, standard deviation and diff. That is, we consider the feature value over the previous six hours and take min, max, mean, median, standard deviation and diff (the standard deviation of the difference of feature value in subsequent hours) and add these as new features. Note that we perform forward fill before this step. So, if there has been a measurement of the specific feature before the current hour, all the values will be non-null. If no measurement of a specific feature has happened before the current hour, then the new feature values are set to $0$. These are motivated from ~\cite{YangWGLLL19}.

\subsection{Classification Algorithm} 
We used XGBoost as the classification algorithm in our pipeline ~\cite{ChenG16}.   It is a gradient boosted tree based algorithm and performs extremely well on structured numerical data. It is an efficient algorithm which beats several of its competitive algorithms as evidence by it's success in data mining contests. It has also been used previously in ~\cite{YangWGLLL19} for Sepsis prediction using only numerical data. Lastly, it handles missing data internally which is an important factor for us since there are many missing entries.


\section{Experiments}

\subsection{Evaluation}\label{sec:utility_score}
\para{Train-validation-test split}
As is common in training and evaluating ML models, we split our dataset into a train, validation and test subset. 
After the data processing steps, we have data for $49168$ patients with $1991$ patients having Sepsis. A total of $1027079$ hours of data is available for these patients. We construct test data containing 7376 patients (299 sceptic, 7077 non-sceptic).
For the remaining patients, we consider five random splits into train set of 33434 patients (1353 septic, 32081 non-septic)  and validation set of 8358 patients (338 septic, 8020 non-septic). As per the data processing and labelling procedure described in Section~\ref{sec:data_processing_labelling}, this results in $9471$ hourly datapoints with Sepsis-positive label and more than 150,000 hourly datapoints with Sepsis-negative label. 
This class imbalance occur mainly due to the fact that only a small percentage of patients develop Sepsis. Also, we use only six hours of data for sceptic patients compared to the entire ICU stay for non-sceptic patients.  Training on such an imbalanced data incentives the model to focus more on correctly identifying non-sceptic rather than correctly identifying sceptic patients, resulting in an almost trivial model making Sepsis-negative predictions for most of the patients. 
To overcome this, we down-sample the negative class such that there are equal number of positive and negative samples in the training set ($9471$ Sepsis-positive hours, $9471$ Sepsis-negative hours). We do not downsample the validation set as it is used to optimize utility score which does not depend on hourly data and instead considers each patient as a datapoint.  We do not downsample the test set as well as it does not affect the training procedure and represents the the actual data observed.

\begin{figure}[ht]
    \centering
    \includegraphics[scale=0.35]{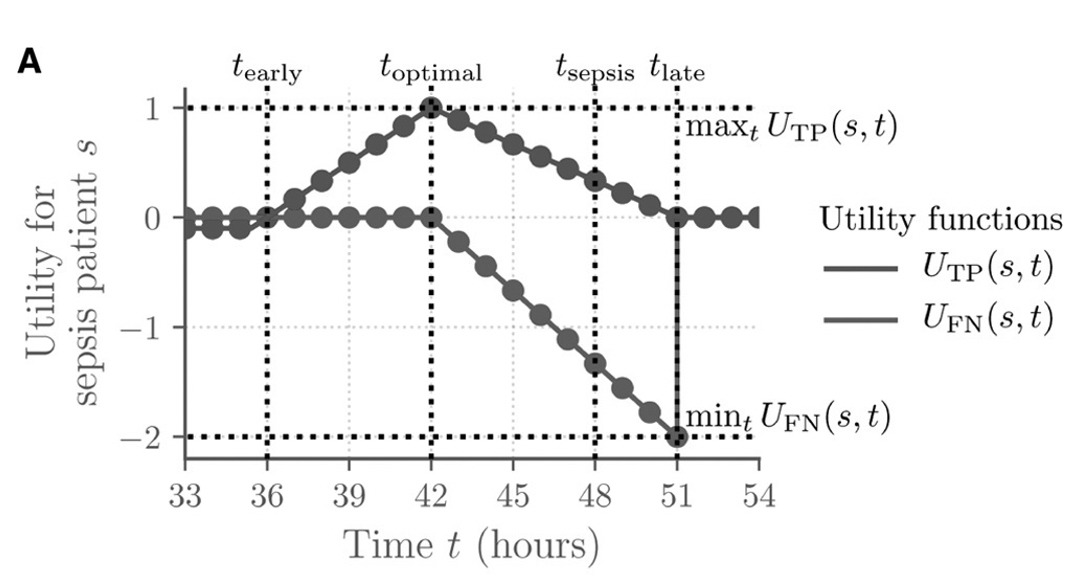}
    \includegraphics[scale=0.35]{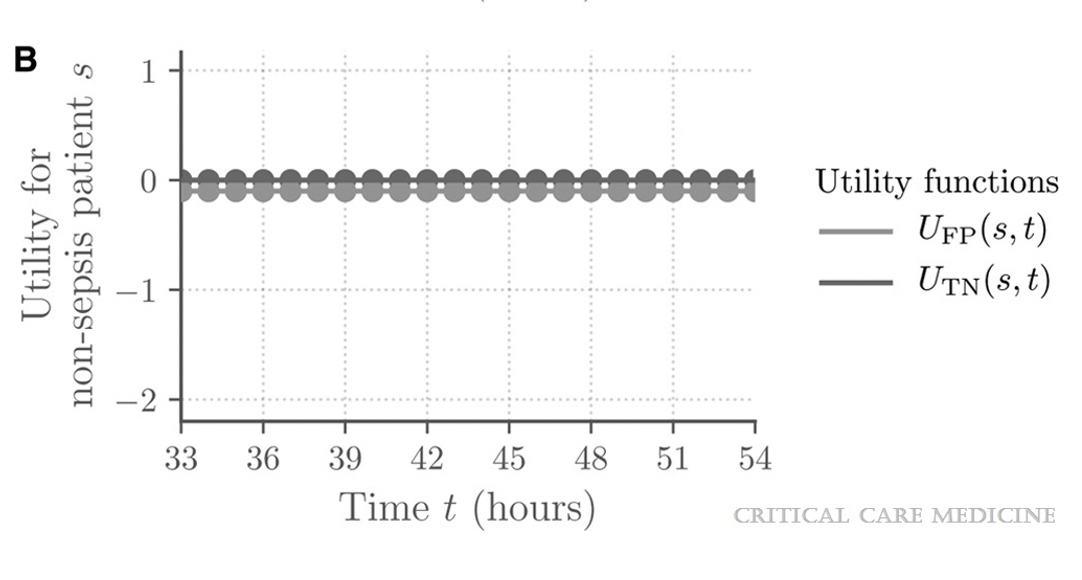}
    \caption{On the left, utility of sepsis prediction (True Positive) and non-sepsis prediction (False Negative) for septic patients. On the right, utility of sepsis prediction (False Positive) and non-Sepsis prediction (True Negative) for non-septic patients. Onset time for Sepsis in these plots is assumed to be 48 hours after admission. Figures are taken from~\cite{SepsisChallenge}.}
    \label{fig:utility}
\end{figure}

\para{Utility Score:} To evaluate and compare performance of different models, the main metric we use is the utility score as defined in The PhysioNet/Computing in Cardiology Challenge 2019 for early Sepsis Prediction~\cite{SepsisChallenge}. 
It assigns positive reward for all the correct Sepsis predictions which happen at most 12 hours before and at most hours 3 hours after the Sepsis onset time. Maximum reward is achieved if the algorithm predict Sepsis roughly six hours before the onset time. Sepsis predictions more than 12 hours before the onset time were slightly penalized as these may be implausible or unhelpful. Non-sepsis predictions for septic patients were increasingly penalized with maximum penalty if the prediction is negative even three hours after the onset time. To account for alarm fatigue, lower algorithm confidence as well as antibiotic overuse, positive Sepsis prediction for non-septic patients were slightly penalized. Non-sepsis prediction for non-septic patient is not rewarded. See Figure~\ref{fig:utility} for exact variation with time.

For each patient ($s \in S$) and for each hour of stay ($t \in T(s)$), we sum up the utility of the predictions to get the score for an algorithm $U = \sum_{s \in S} \sum_{t \in T(s)} U(s,t)$.
Then, we normalize the score such that the optimal algorithm receives a reward of $100$ and the algorithm making no sepsis predictions receives a reward of $0$, $U_{normalized} = 100\times\frac{U_{total} - U_{no-prediction}}{U_{optimal} - U_{no-prediction}}$

\para{Other metrics} we use to evaluate the different methods are more standard to binary classification tasks. For every method we report the  area under the receiver operating characteristic ({\bf AUROC}) score. It is equal to the probability of the trained model producing higher risk score for a randomly selected sceptic patient than a randomly selected non-septic patient. It lies  in the range from $0$ to $1$. A model producing random numbers will on average achieve an AUROC score of $0.5$, and a model making perfect predictions will obtain a score of 1. It is based on the raw scores produced by the model instead of the sepsis/no-sepsis label after thresholding. We report the AUROC score for the hourly datapoints in the test data.  
%
In addition to AUROC score, we also report the false positive rate ({\bf FPR}) defined as the percentage of non-septic patients identified as septic, and the true positive rate for septic patients $x$ hours before Sepsis onset ({\bf TPR-$x$}), meaning the percentage of septic patients that will be identified to be at risk at least $x$ hours before Sepsis onset.

\subsection{Baseline Methods}\label{subsec:baselines}

In addition to the methods detailed in Section~\ref{sec:text}, we evaluate the following baselines:

\para{Numerical:} This refers to the model where we ignore all the textual features. We keep the pipeline described in Figure~\ref{fig:alg_pipeline}
but we do not include any features corresponding to the text. This is very similar to the pipeline used in \cite{YangWGLLL19} which won the PhysioNet Computing in Cardiology Challenge 2019 for predicting Sepsis \cite{SepsisChallenge}. \cite{YangWGLLL19} uses a different set of features and performs some hyperparamter optimization steps but uses the same data labeling, data processing, feature augmentation and the binary classifier.

\para{qSOFA:} Recall that many of the values taken for measuring the SOFA score in defining sepsis are taken in labs, which are taken infrequently and require a long delay. In addition to define Sepsis-3, the authors, in~\cite{SingerDSSABBBCC16} who are clinical experts,  introduced a quick method to identify Sepsis-3 using the data that can be easily collected. It is primarily designed for hospital stay but can be applied to ICU stay as well. This is based on satisfying any 2 of 3 criteria: 1) having a significant change in mentation - represented by a Glasgow comma score of 12 or less, 2) respiratory rate $\geq$ 22/min, or 3) systolic blood pressure $\leq$ 100 mmHG. This foerms another baseline method with which we compare our method.

\para{tf-idf:} In this model, we use the the classic term frequency-inverse document frequency (tfidf) based embedding for incorporating text instead of contextualized embeddings given by BERT. In tfidf embedding, the value increases proportionally to the number of times a word appears in the document and is offset by the number of documents containing that word. We merge all the textual features and build a tfidf embedding based on a vocabulary of the $1000$ most common words resulting in a $1000$ dimensional embedding. Rest of the pipeline is same as in Figure~\ref{fig:alg_pipeline}.


\subsection{Results}

\begin{table}[ht]
    \centering
    \setlength{\tabcolsep}{3.2pt}
    \footnotesize
    \begin{tabular}{cccccccccccc}
    \toprule
    Method & Model & Utility Score & AUROC & FPR & \multicolumn{7}{c}{TPR $x$ hours before onset}\\
   & & & &  & $x=6$ & $x=5$ & $x=4$ & $x=3$ & $x=2$ & $x=1$ & $x=0$\\
    \midrule
    &Numerical & $42.73_{\scriptscriptstyle{0.95}}$ & $86.42_{\scriptscriptstyle{0.32}}$& $27.54$& $70.10$	&$74.58$&	$77.12$&	$79.26$	&$81.00$&	$81.94$	&$83.01$\\
   Baselines &tfidf & $44.83_{\scriptscriptstyle{0.85}}$ &$87.22_{\scriptscriptstyle{0.21}}$ & 24.61&72.91&76.86& 80.33&82.74&83.81&84.41&85.35
\\
   & qSOFA & $-23.47_{\scriptscriptstyle{0}}$ & $56.64_{\scriptscriptstyle{0}}$ &72.66&70.23&71.24&	72.91&73.24&74.92&75.59&80.27\\
   \midrule
   \midrule
    Comprehend&CMBERT & $43.25_{\scriptscriptstyle{0.52}}$ & $86.34_{\scriptscriptstyle{0.28}}$& 29.07&73.11&77.39&80.67&82.27&83.95&84.88&85.95\\
   Medical &CMtfidf & $44.73_{\scriptscriptstyle{0.56}}$ & $87.32_{\scriptscriptstyle{0.37}}$& 25.88&72.31&76.66&79.87&82.14&83.21&83.55&84.75\\
    \midrule
    \midrule
    &BERTM & $42.93_{\scriptscriptstyle{0.36}}$& $86.24_{\scriptscriptstyle{0.32}}$&26.53	&70.64	&74.58	&78.39	&81.00	&82.47	&83.41	&84.35\\
    BERT&BERTS & $43.24_{\scriptscriptstyle{0.51}}$&$86.51_{\scriptscriptstyle{0.02}}$ & 28.33&72.64&	76.86&79.00&81.14&82.21&83.28&84.28\\
    &FBERTM & $47.55_{\scriptscriptstyle{0.38}}$& ${87.77_{\scriptscriptstyle{0.39}}}$& 24.89&74.31&78.73&81.27&84.08&85.08&86.22&87.22\\
    &FBERTS & ${\bf 48.80_{\scriptscriptstyle{0.29}}}$&${\bf 89.31_{\scriptscriptstyle{0.27}}}$ & 23.45&73.58&78.73&81.14&83.95&84.95&85.82&86.62\\
\end{tabular}
    \caption{Numerical, tfidf, and qSOFA refers to the baseline models as discussed in Section~\ref{subsec:baselines}. CMBERT, CMtfidf, BERTM, BERTS, FBERTM, FBERTS refer to the text embedding techniques described in Section~\ref{sec:text}.
    Five models are trained on train data and thresholds are selected based on validation data. Utility score of $0.45_{\scriptscriptstyle{0.1}}$ refers to the average utility score of $0.45$ with a standard deviation of $0.1$ of the five trained models on the test data. Similarly, AUROC score of $85.92_{\scriptscriptstyle{0.31}}$ refers to the average AUROC score of $85.92$ with a standard deviation $0.31$. FPR stands for false positive rate, meaning the percentage of non-septic patients identified as septic. TPR stands for true positive rate and is measured for septic patients at different times before septic onset.}

    \label{tab:results}
\end{table}

Table~\ref{tab:results} shows the results of our models as well as several baseline models. 
For each of the entry, we train five models with different train/validation split and report the mean and standard deviation of the performance on the test set. Note that AUROC score is computed w.r.t. the hourly datapoints in the test set whereas rest of the metrics are computed by considering each patient in the test set as a datapoint.  Each of the trained model produce raw scores which are converted into Sepsis/No-Sepsis prediction based on thresholding (for all metrics except AUROC). In our results, thresholds are chosen so as to maximize utility score on the validation set with the exception of {\bf FBERTM} and {\bf FBERTS} which use the default 0.5 cutoff.
This is because {\bf FBERTM} and {\bf FBERTS} contain outputs from a BERT-based model trained on the entire training set so the validation sets are not representative of the final test set. We can modify thresholds if our objective is to optimize a different quantity such as accuracy.

We note that our model {\bf FBERTS} significantly outperforms the baseline numerical model not utilizing the text. Incorporating text can boost the utility score by roughly $6$ points and AUROC score by roughly $3\%$. The FPR of {\bf FBERTS} is $4\%$ less than that of {\bf Numerical}. That is, for non-septic patients our model predicts sepsis in $4\%$ less cases than {\bf Numerical}. Simultaniosly, the TPR of {\bf FBERTS} for $3$ hours before onset is $4.69\%$ higher than that of {\bf Numerical}. That is, for septic patients, our model predict sepsis at least $3$ hours before the onset in $4.69\%$ more patients than {\bf Numerical}. Our model significantly outperforms a clinical practitioner suggested method, qSOFA. {\bf qSOFA} in fact achieves a negative utility score and very low AUROC score. Lastly, to emphasize the importance of the technique of adding text features, we note that {\bf FBERTS} outperforms the {\bf tfidf} by roughly $4$ points in utility score and roughly $3\%$ in AUROC score. 

We also note that the model incorporating Comprehend Medical ({\bf CMBERT} and {\bf CMtfidf}) performs worse than the the model incorporating BERT embeddings. In fact, tfidf embeddings on the extracted entities by Comprehend Medical ({\bf CMtfidf}) performs better than taking BERT embeddings on the extracted entities concatenated together. This is likely due of the fact that BERT is inherently designed to generate contextualized embeddings taking into account properties related to sentence structure and relationships between words. Since text with extracted entities is not a regular text with appropriate grammatical structure, BERT fails to produce good contextual embeddings on the extracted entities. 
There may be ways to mitigate this, but given the success we observed with {\bf FBERTS}, we leave this for future work. 


Note that we report our results even 0 hours before Sepsis onset time. Predictions make sense even in this time frame because we do not always get the Sepsis diagnosis at the same moment as when the Sepsis occur (as per definition in Section~\ref{subsec:sepsis_definition}). This is because, there may be a lag in the lab results and they may take several hours to run and return. For instance, if a lab test is ordered at 4 P.M. whose result at 6 P.M. concludes that the patient has Sepsis, then the Sepsis onset time is defined as 4P.M. However, the diagnosis can only be made at 6 P.M. Also, note that we have a relatively high false positive rate. For roughly $30\%$ of the non-septic patients, we predict Sepsis. This is desirable compared to high false negative rate as the cost of false negative significantly outweigh the cost for false positive prediction. It is also reflected in the utility score as the reward for false negative is significantly more negative than the reward for false positive. 





\begin{figure}[ht]
    \centering
    \includegraphics[scale=0.18]{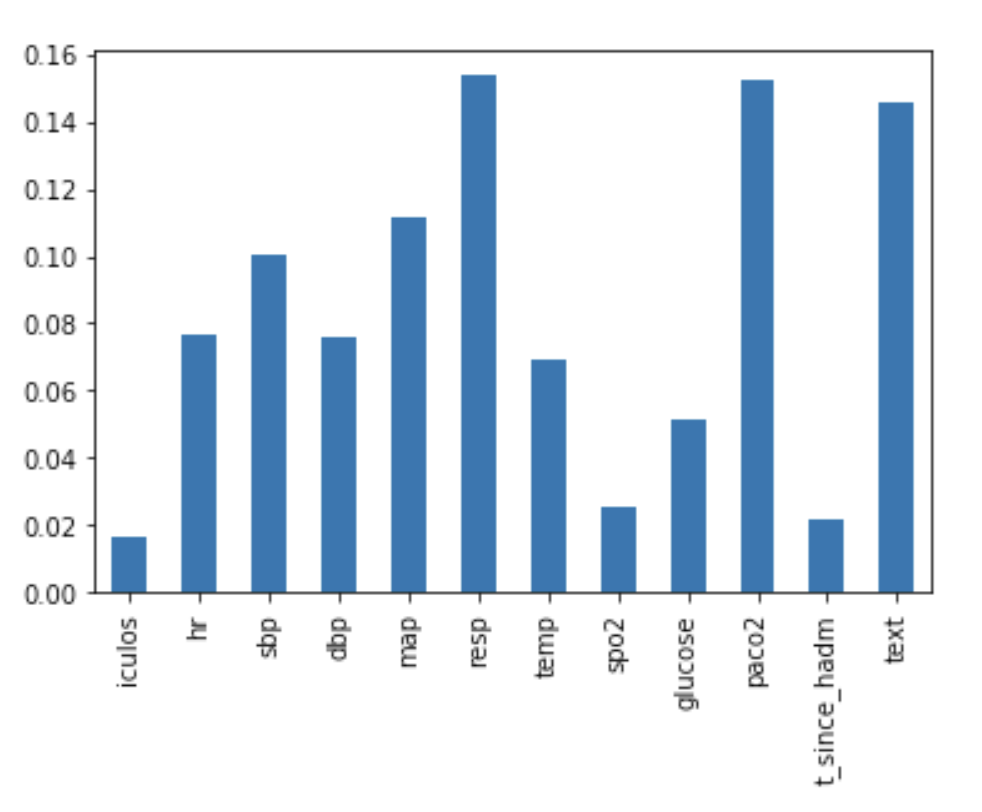}
    \caption{Feature Importance of different features in the trained FBERTS model.}
    \label{fig:feature_importance}
\end{figure}

\para{Feature importance:}
After training the model, we used Shapley importance score~\cite{lundberg2020local} to evaluate the importance of each of the original features used. A higher score for one feature vs.\ the other means it is more influential in determining the final score of the model.
The XGBoost model provides such scores for its input (augmented) features. For a numerical feature such as hr, we sum up the feature importance score of all the augmented features coming from hr. For textual features, we sum up the scores of all textual features. Figure~\ref{fig:feature_importance} presents the obtained results. We see that the text features are the third most influential in determining Sepsis onset. 

\section{Related works}

\para{Models for Sepsis Prediction:} There have been a series of  works on applying machine learning techniques for sepsis prediction~\cite{DesautelsCHJKSSCFB16,HorngSHJSN17,NematiHRSCB18,MaoJHCBSSCFK18,CalvertPCBFHJD16,KamK17,FaisalSRBHSM18}. However, as mentioned above all of these works have been focused on utilizing only structured information such as heart rate, glucose level, etc. Also, several of these papers work with an outdated definition of Sepsis. ~\cite{IslamNWWYL19} perform a comparative study of these works in terms of pooled area under receiving operating curve (SAUROC) for predicting sepsis onset 3 to 4 h before, sensitivity and specificity. It also shows that machine learning methods achieve better performance than the non-machine learning scoring systems for predicting sepsis. 

\para{Models incorporating text for related problems:} There have been a few papers dealing with textual data for related problems. Based on our knowledge, there has only been one sepsis-related study incorporating text that only predicted severe sepsis using textual features by representing clinical notes as 300-dimensional GloVe vectors only, compared it with a model trained on structured 12 numerical features and found that the text-based model performed better in their study than the 12 numerical features \cite{culliton2017predicting}. In addition, several other studies have shown the success of utilizing unstructured text data along with structured data to predict disease: \cite{wang2015early} used a combination of structured and unstructured data for early prediction of heart failure, 
~\cite{LiuZR18} used embeddings from CNNs and BiLSTM to represent text for predicting chronic diseases, and ~\cite{JinBCBCBSKNZ18} used Doc2Vec algorithm to embed notes in mortality prediction. 

\section{Conclusion}
We observe that incorporating textual features such as nurse's and doctor's notes can significantly boost the performance of Sepsis prediction model. We present a BERT based model to incorporate text which achieves this improvement and beats the competing method not utilizing text \cite{YangWGLLL19} which won the PhysioNet Computing in Cardiology Challenge 2019 for predicting Sepsis. It also beats the baselines methods including a clinical practitioner suggested method qSOFA and a naive tfidf embedding based method to incorporate text.  
Our model can be easily adapted for other tasks such as morbidity or mortality prediction, or predictive modeling of hospital readmission. We can improve our model by further exploration into text embedding procedure. For instance, entities from ComprehendMedical can be filtered and we may only use entities such as medications or medical conditions as they are more likely to be useful for Sepsis predictions.

\bibliography{arxiv}
\bibliographystyle{plain}
\end{document}